%% file: neurips_2026.tex
\documentclass{article}

\usepackage[main,final,nonatbib]{neurips_2026}

\usepackage[utf8]{inputenc} %
\usepackage[T1]{fontenc}    %
\usepackage[hidelinks]{hyperref}       %
\usepackage{url}            %
\usepackage{booktabs}       %
\usepackage{amsmath,amsthm,amsfonts,amssymb}       %
\usepackage{nicefrac}       %
\usepackage{microtype}      %
\usepackage{xcolor}         %
\usepackage{algorithm}
\usepackage{subcaption}
\usepackage{algpseudocode}
\def\shownotes{1}

\usepackage{macro}

\usepackage[backend=biber,
            natbib=true,
            maxbibnames=99,
            minalphanames=3,
            uniquename=false,
            uniquelist=false,            
            maxcitenames=2,
            sorting=ynt,
            giveninits=true,                        
            labeldateparts=true,
            sortcites=true,
            backref=false]{biblatex}
\title{Structured Kernel Surrogate Models for\\Multimodal Attribution}

\author{%
  Ziniu Zhang\thanks{Equal contribution.}\\
  Northeastern University\\
  Boston, MA\\
  \And  
  Zhenshuo Zhang\footnotemark[1]\\
  Northeastern University\\
  Boston, MA\\
  \And  
  Jianglin Lu\\
  Northeastern University\\
  Boston, MA\\
  \And
  Ruoxuan Xiong\\
  Emory University\\
  Atlanta, GA\\
  \And
  Yun Fu\\
  Northeastern University\\
  Boston, MA\\
  \And
  Hongyang R. Zhang\\
  Northeastern University\\
  Boston, MA\\
}

\begin{document}

\maketitle

\input{intro.tex}
\input{content.tex}

\input{experiments.tex}

\input{related}

\begin{refcontext}[sorting=nyt]
    \printbibliography
\end{refcontext}

\appendix

\newpage
\input{appendix}

\end{document}

%% file: intro.tex
\begin{abstract}
    We study the problem of \emph{cross-sample modality attribution}, which involves measuring how one modality in one sample affects the learning outcome in other multimodal samples. This problem is crucial in a multimodal system because a data modality (such as a text description) may be redundant for one sample (a self-explanatory image), but the same data modality would otherwise be uniquely informative for another sample, or informative when combined with other modalities. Prior work focuses on influence estimation at the level of one sample, and it remains an open question how one should capture the interactions across both samples and modalities. To capture scenarios like the above, we propose \emph{structured kernel surrogate} models, which are surrogate mappings from both samples and modalities to the prediction outcome when they are trained together. We design an algorithm, \algo{}, which applies kernel ridge regression with a structured, compositional kernel on the selection states of individual samples and modalities. Specifically, the kernel function multiplies kernel-based similarity scores between every pair of data modalities for one sample and a pair of two samples. Additionally, we design a gradient-based estimation algorithm to efficiently estimate the structured kernel surrogate model. We evaluate our approach on two multimodal benchmarks, spanning a range of data modalities including satellite images, road networks, text, audio, and tabular data. We find that \algo{} achieves $57.8\%$ higher correlation with the true prediction outcomes than existing methods. In downstream data selection, \algo{} improves the average accuracy by $1.93\%$ over baselines. Models with data selected by \algo{} have lower Hessian traces, indicating a flatter loss surface. Using \algo{} to select satellite images identifies road segments, where accident scenes commonly occur, as opposed to background images.
    \end{abstract}
    
    \section{Introduction}\label{sec_intro}
    
    Data in the form of multiple modalities are widely available in many domains. For example, in transportation, satellite images capturing road segments, traffic statistics such as traffic volume, and weather statistics can jointly be used to make accurate predictions on accident risks \cite{nippani2023graph,zhang2026learning}. Autonomous self-driving vehicles integrate visual cues and sounds from the surrounding environment to provide safety-critical navigation \cite{balasingam2024drivetrack}. While multimodal systems have scaled up in terms of both model size and complexity, their internal mechanisms remain highly complex \cite{zhang2026efficient} (one concrete example where susceptible models may be brittle to feature representations is spurious correlations \cite{pmlr-v162-zhang22z}). Thus, understanding how different modalities interact and impact predictions is central to providing meaningful interpretations to multimodal systems.
    
    Compared to a unimodal system, data influences a multimodal system at different granularities. In this vein, the problem of cross-sample modality interaction can be related to earlier literature on multi-view learning. For instance, \citet{kakade2007multi} use relationships estimated from unlabeled paired data to regularize predictors and encourage agreement across views (see also a follow-up work \cite{sridharan2008information}).
    Another related concept built on partial information decomposition characterizes redundant, unique, or synergistic information in the presence of heterogeneous data modalities, under different combinations~\cite{liang2023quantifying,gupta2025better}. 
    While these existing works describe what information is present across modalities,
    the question of how individual samples and their modalities affect prediction remains open: is there an efficient algorithm for measuring the effect of adding or removing one modality from an individual sample on model performance?
    
    Prior work in the literature of influence estimation studies the effect of adding or removing an individual example on model predictions. Existing methods, including influence functions~\cite{koh2017understanding}, datamodels~\cite{park2023trak}, and linear and kernel surrogate methods~\cite{liidentification,zhang2026efficient}, have considered the estimation problem both for individual samples and for groups (or sets) of samples. These existing works primarily focus on unimodal settings. In particular, the effect of adding or removing one modality from a sample can depend on how it interacts with other samples and modalities through a shared network or optimization procedure. The problem of estimating the influence of individual samples in multimodal learning remains open.
    
    \begin{figure}[t]
        \centering
        \includegraphics[width=1.00\linewidth]{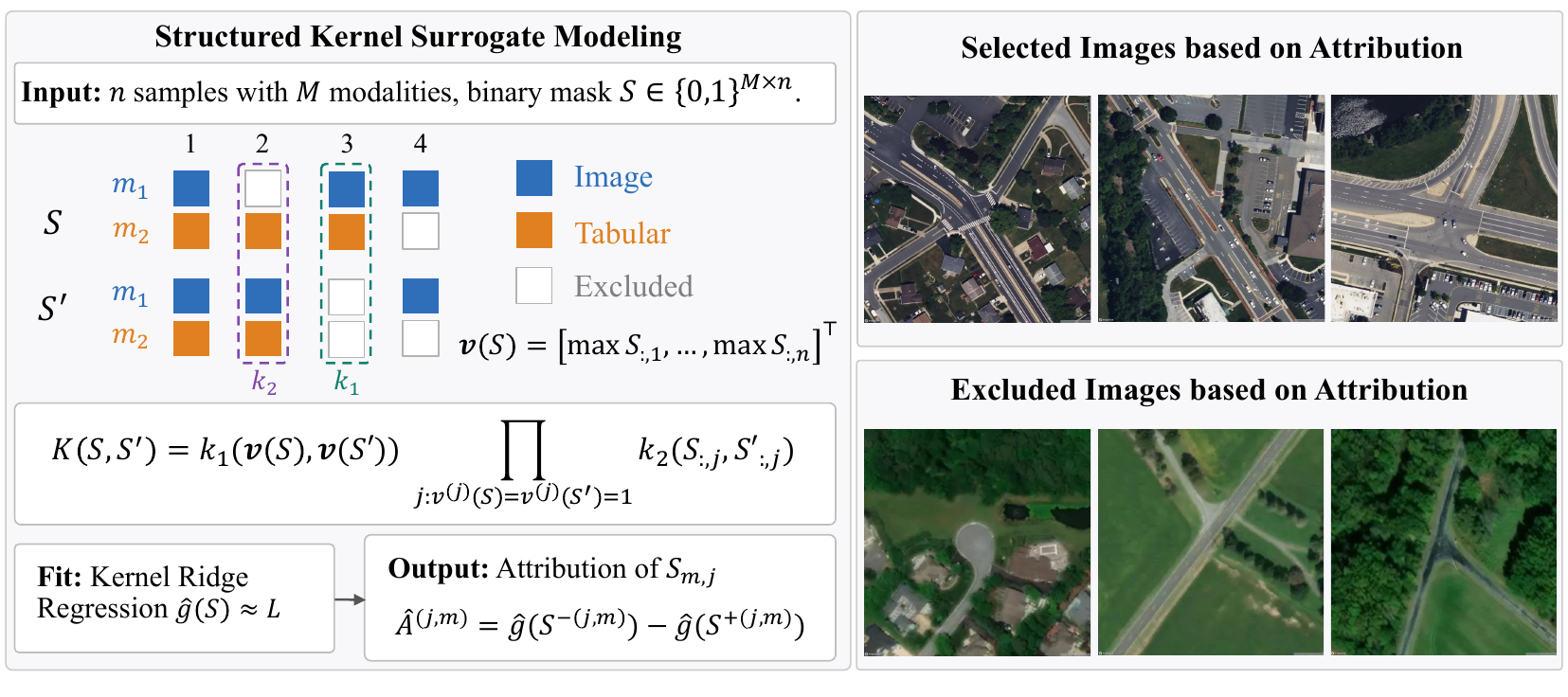}
        \caption{Overview of our proposed approach. We design a structured compositional kernel by multiplying both sample-level and modality-level RBF kernel similarity scores, given by $k_1$ and $k_2$ (with bandwidth parameters $\gamma_1$ and $\gamma_2$, respectively). Then, we fit a surrogate model mapping the selection states to their loss outcomes using kernel ridge regression on $N$ pairs of randomly sampled selections and their corresponding loss outcomes (after fitting a model). Lastly, we estimate the influence score of one sample-modality pair $(m ,j)$ under a binary mask $S_{m,j}$ by subtracting the test losses under $S^{-(j,m)}$ and $S^{+(j,m)}$, which retain and remove this modality in the sample, with all other entries of $S$ fixed. A positive score indicates that removing this pair $(m, j)$ is detrimental. On the right, we give examples of satellite images retained and excluded based on their influence scores. Our approach correctly identifies the road segments from the top panel, where accident scenes often occur, and estimates a higher influence score for these segments compared to the background images from the bottom panel.
        }\label{fig_algorithm}
    \end{figure}
    
    These gaps motivate us to introduce \emph{structured kernel surrogate} models, or \algo{} for short, for capturing cross-sample modality interaction with modality missingness and nonlinear relationships. The entire framework is illustrated in Figure \ref{fig_algorithm}. \algo{} considers perturbations of the dataset in which some samples are included or removed, and some modalities are retained or masked within each included sample. For each such sampling, the corresponding target is the loss of the model based on the perturbed dataset. \algo{} then learns this selection-to-loss map using kernel ridge regression with a structured radial basis function (RBF) kernel. The kernel separates sample-level participation from within-sample modality composition, which crucially allows for missing data modalities. Moreover, it captures nonlinear interactions across samples and modalities. An implication of this work is thus connecting the literature on influence estimation in multimodal learning to concepts in causal inference (such as difference-in-differences and modality missingness \cite{liang2025causal,liang2026learning}).
    
    We evaluate \algo{} on the multimodal traffic accident prediction benchmark \textsc{MMTraCE} \cite{zhang2026learning} and the standard multimodal benchmark MultiBench \cite{liang2021multibench}. We first measure attribution quality using the linear data modeling score (LDS). Compared with the strongest baseline, \algo{} improves the per-task relative LDS by $50.1\%$ on \textsc{MMTraCE} and $65.6\%$ on MultiBench. We then evaluate whether these attributions can guide downstream data selection. Under a fixed top-$20\%$ vision-modality selection budget, \algo{} improves the average accuracy by $1.52\%$ on \textsc{MMTraCE} and by $2.33\%$ on MultiBench over the strongest baseline. Similar gains are observed when the selected modality is changed to tabular information. Finally, we analyze the generalization behavior of models from data selected by \algo{}. The selected data lead to strong cross-state generalization and a smoother loss surface than competing attribution methods. This shows better robustness of the resulting models using \algo{}.
    
    In summary, our contributions are threefold. First, we propose \algo{}, a \emph{structured kernel surrogate} model that separates sample-level and modality-level selection to learn the nonlinear subset-to-loss mapping. We provide a sample complexity bound for the surrogate.
    Second, \algo{} achieves the highest LDS across six MultiBench tasks, improving over the strongest baseline by $65.6\%$ in per-task relative LDS on average, with the largest gains on MOSI and Sarcasm. It also improves downstream test accuracy on MultiBench by $2.33\%$ on average.
    Third, on spatiotemporal traffic accident prediction across six US states, \algo{} improves LDS by $50.1\%$ over the strongest baseline. For downstream data selection, it improves average test accuracy by $1.52\%$. The learned attributions are also interpretable: \algo{} assigns higher scores to satellite images with visible road infrastructure than to vegetation-dominated tiles. Models trained on data selected by \algo{} show strong cross-state transferability. The loss sharpness analysis further shows that they achieve the lowest Hessian trace against baselines.
    The code repository for reproducing the empirical findings can be found at \href{https://github.com/VirtuosoResearch/Multimodal-attribution-SKS-code}{https://github.com/VirtuosoResearch/Multimodal-attribution-SKS-code}.

%% file: content.tex
\section{Preliminaries}\label{sec_pre}

Let $D=\{(\mathbf{x}_i,y_i)\}_{i=1}^n$ denote a multimodal dataset sampled from an unknown distribution $\cD$, where $\mathbf{x}_i$ is the multimodal input with $M$ modalities, and $y_i$ is the output. For exposition, we focus on the two-modality setting $\mathbf{x}_i=(x_i^{(1)},x_i^{(2)})$. In traffic accident prediction, for example, each sample may represent a road segment, modeled as an edge in a road network graph. The two modalities can consist of a satellite image of the segment and tabular features such as traffic volume.
We learn a multimodal network $f_W$ on selected samples and selected modalities within each sample. The global selection state is represented by the binary matrix
\[
{S}
=
\begin{bmatrix}
S_{1,1} & S_{1,2} & \cdots & S_{1,n}\\
S_{2,1} & S_{2,2} & \cdots & S_{2,n}
\end{bmatrix}
\in \{0,1\}^{2\times n},
\]
where $S_{m,i}=1$ means that modality $m$ of sample $i$ is retained and $S_{m,i}=0$ means that it is removed. Thus, column $i$ of $S$ indicates which modalities are retained in sample $i$, and row $m$ indicates which samples retain modality $m$.  The extension to more than two modalities is straightforward and is discussed in detail in Appendix~\ref{app_extension}. We define 
\[ \mathbf{v}(S)=[\max(S_{:,1}),...,\max(S_{:,n})]^\top. \]
In particular, $\mathbf{v}^{(i)}(S)=1$ if at least one modality of sample $i$ is retained, and $\mathbf{v}^{(i)}(S)=0$ if the entire sample is removed. 

For sample $i$, the masked loss is denoted as $\ell\big(f_W(\mathbf{x}_i;S_{:,i}),y_i\big)$. We require at least one sample to retain at least one modality, so $\sum_{j=1}^n \mathbf{v}^{(j)}(S)>0$. We define the empirical loss under $S$ as the average loss over all samples: 
\[  
\hat{L}(f_W; S) = \left(\sum_{j=1}^n \mathbf{v}^{(j)}(S)\right)^{-1} \sum_{i=1}^n \mathbf{v}^{(i)}(S) \, \ell(f_W(\mathbf{x}_i; S_{:,i}),y_i). 
\]
The learned parameters $W$ generally depend nonlinearly on $S$. Changing a modality in one sample changes the corresponding sample's contribution to the objective and, through shared parameter updates, can further affect how the model uses information from other samples. 

We characterize the effect of a selection ${S}$ by the expected risk of the resulting model under the test distribution $\mathcal D$:
\[ 
L(f_{W};S) = \mathbb{E}_{(\mathbf{x},y)\sim \mathcal D}\big[\ell(f_{W}(\mathbf{x};S),y)\big]. 
\]
Our goal is to characterize how perturbing a single entry of $S$, corresponding to retaining or removing one modality of one sample, affects the resulting test loss $L(f_W;S)$.

\textbf{Problem statement.} 
We introduce an attribution score to measure how one modality of an individual sample affects the test loss. For modality $m$ in sample $j$, let $S^{+(j,m)}$ and $S^{-(j,m)}$ be obtained from $S$ by setting $S_{m,j}$ to $1$ and $0$, respectively, keeping all other entries unchanged. We define the influence score of modality $m$ in sample $j$ at ${S}$ as 
\begin{equation}
    A^{(j,m)}({S}) = L(f_W;{S}^{-(j,m)}) - L(f_W;{S}^{+(j,m)}).\label{eq:modality_attribution}
\end{equation}
$A^{(j,m)}({S})$ measures the change in test loss caused by removing modality $m$ from sample $j$ and keeping other sample and modality choices unchanged. A positive value indicates that retaining the modality in this sample reduces the test loss. We note that this is dependent on the binary mask matrix $S$. This formulation raises two questions:
\begin{itemize}
    \item First, how can we estimate the nonlinear mapping from the binary masks of samples and modalities specified by $S$ to the loss outcome $L(f_W; S)$?
    \item Second, how can we use the surrogate model to efficiently estimate the influence score for each modality of an individual sample?
\end{itemize}
The next section proposes a detailed methodology to answer the two questions.

\section{Our Approach}\label{sec_main}

We address the two questions in turn. 
We first introduce a structured kernel surrogate to model the mapping from $S$ to  $L(f_W;S)$, with a structured similarity that separately accounts for sample participation and modality composition. 
We then recover the attribution scores through finite differences of the fitted surrogate.

\subsection{Designing Kernel Surrogate Models}\label{sec_srbf}
To estimate the mapping from $S$ to  $L(f_W;S)$, we model how individual samples and their modalities interact and jointly affect the test loss. These interactions arise at two levels: sample-level interaction and modality-level interaction. We model these two types of interactions separately using two RBF kernels. We then combine the two kernels to form a structured design that captures interactions among samples and modalities. We utilize the standard RBF kernel with squared Euclidean distance.\footnote{We provide both a justification for the RBF kernel and an ablation of alternative kernel choices in Appendix~\ref{app_kernel_choice}.}

\paragraph{A compositional kernel.} 
Given two binary mask matrices $S, S^{\prime}$, we first quantify their sample-level similarity by comparing $\mathbf v(S)$ and $\mathbf v(S')$ using an RBF kernel with bandwidth $\gamma_1>0$:
\[ 
k_{1}(\mathbf{v}(S), \mathbf{v}(S^{\prime})) = \exp \left(-\gamma_1 \|\mathbf{v}(S) - \mathbf{v}(S^{\prime})\|_2^2 \right). 
\]
Since $\mathbf{v}(S), \mathbf{v}(S^{\prime})$ are binary, the squared distance counts the samples whose participation differs between the two states.

For each sample $j$ that is selected in both $S$ and $S'$, we compare $S_{:,j}$ and $S'_{:,j}$ to measure how similarly the modalities of this sample are selected across the two states. We define this modality-level similarity using an RBF kernel with bandwidth $\gamma_2>0$:
\[ 
k_{2}(S_{:,j}, S'_{:,j}) = \exp \left(-\gamma_2 \|S_{:,j} - S'_{:,j}\|_2^2\right). 
\]

We then combine the sample-level and modality-level kernel similarity scores to define the structured similarity between $S$ and $S'$:
\[
K(S, S')
= k_1(\mathbf v(S), \mathbf v(S'))
\cdot \left(
\prod_{j\in[n]:\,\mathbf{v}^{(j)}(S)=\mathbf{v}^{(j)}(S')=1}
k_2(S_{:,j}, S'_{:,j})
\right).
\]

Restricting the product to samples that are present in both $S$ and $S'$ separates sample removal from modality changes within a retained sample. If a sample is absent in either $S$ or $S'$, its difference is captured by the sample-level component $k_1$ and does not contribute an additional modality-level term.

This design allows $K$ to capture interactions between samples. It also captures interactions between modalities within a sample and across samples. For two modalities, $K$ is positive semidefinite if $\gamma_2>0$ and $\gamma_1 \ge \frac{1}{2}\log\left(\frac{3-e^{-\gamma_2}}{1+e^{-\gamma_2}}\right)$. Under these conditions, $K$ is a valid kernel. We give the proposition and proof in Appendix~\ref{app_psd}.

\begin{algorithm}[t!]
\caption{Estimating Structured Kernel Surrogate (\algo{}) Models for Multimodal Attribution}
\label{alg_srbf}
\textbf{Input:} $N$ randomly sampled selections $\set{S_1, S_2, \dots, S_N}$ and their evaluation loss outcomes based on corresponding selections $\{(S_p, \tilde L(f_W; S_p))\}_{p=1}^N$\\
\textbf{Require:} 
Sample-level RBF bandwidth $\gamma_1$, modality-level RBF bandwidth $\gamma_2$, reference selection matrix $S$, ridge regression penalty $\lambda$\\
\textbf{Output:} Surrogate model and estimated influence scores %
\begin{algorithmic}[1]
    \Statex \textit{/* Stage 1: calculate the kernel and solve a kernel ridge regression problem \hfill */}
    \For{$p,q=1,\ldots,N$}
        \State $\mathcal{J}_{pq}
        \leftarrow
        \{j\in[n]: \mathbf v^{(j)}(S_p) = \mathbf v^{(j)}(S_q) = 1\}$
        \State $\mathbf{K}_{pq}
        \leftarrow
        k_1\!\left(
            \mathbf{v}(S_p),
            \mathbf{v}(S_q)
        \right)
        \prod_{j\in\mathcal{J}_{pq}}
        k_2\!\left(
            S^{(p)}_{:,j},
            S^{(q)}_{:,j}
        \right)$
    \EndFor
    \State $\mathbf{y} \leftarrow (\tilde L(f_W;S_1), \ldots, \tilde L(f_W;S_N))^\top$
    \State $k_{S} \leftarrow (K(S,S_1),\ldots,K(S,S_N))^\top$
    \State $\hat{g}(S) \leftarrow k_{S}^\top (\mathbf{K} + N \lambda \mathbf{I})^{-1} \mathbf{y}$
    \Statex \textit{/* Stage 2: calculate the estimated influence scores \hfill */}
    \For{each $j\in [n]$ and $m\in[M]$}
        \State $S^{+(j, m)}, S^{-(j, m)} \leftarrow$ modify $S$ by setting $S_{m,j} = 1$ and $0$ respectively 
        \State $\widehat{A}^{(j,m)}(S) \leftarrow \hat{g}(S^{-(j,m)}) - \hat{g}(S^{+(j,m)})$
    \EndFor
    \State \Return $\hat{g}$, $\{\widehat A^{(j, m)}(S) : j\in[n],\; m\in[M]\}$
\end{algorithmic}
\end{algorithm}

\subsection{Surrogate Modeling of Sample-Modality Interaction}

Given a valid kernel, we then fit a surrogate model that predicts the test loss from $S$. We then estimate the attribution of each modality in each sample using finite differences. Each difference compares the predicted losses with the modality added and removed, where all other entries of $S$ remain fixed.

\paragraph{Surrogate modeling.}
Let $\mathcal{H}_K$ denote the reproducing kernel Hilbert space (RKHS) induced by the structured kernel $K$. Given $N$ sampled selection states $S_1, \ldots, S_N$ with measured test losses $L(f_W;S_1),\ldots,L(f_W;S_N)$, we fit a kernel ridge regression surrogate~\cite{zhang2026efficient} by optimizing over candidate functions $g\in\mathcal{H}_K$, where $g(S)$ predicts the test loss associated with state $S$:
\[
\hat{g} = \arg\min_{g \in \mathcal{H}_K} \frac{1}{N} \sum_{p=1}^N (g(S_p) - L(f_W;S_p))^2 + \lambda \|g\|_{\mathcal{H}_K}^2.
\]
Let $\mathbf{K}\in\mathbb{R}^{N\times N}$ denote the Gram matrix over the $N$ sampled selection states, whose $(p,q)$-th entry is $\mathbf{K}_{p q}=K(S_p,S_q)$. Let $\mathbf{y}=(L(f_W;S_1),\ldots,L(f_W;S_N))^\top$ and $k_{S}=(K(S,S_1),\ldots,K(S,S_N))^\top$. The closed-form solution is given by
\[
\hat{g}(S) = k_{S}^\top (\mathbf{K} + N \lambda \mathbf{I})^{-1} \mathbf{y}.
\]

\paragraph{Influence estimation.} Given a reference selection state $S$, we estimate the attribution using the fitted surrogate $\hat g$. For modality $m$ in sample $j$, 
the estimated attribution is 
\[ \widehat A^{(j, m)}(S) = \hat g\!\left(S^{-(j,m)}\right) - \hat g\!\left(S^{+(j,m)}\right).
\]
We compute the finite difference with all other entries fixed at their values in $S$. The attribution depends on which other samples and modalities are retained, and it can reflect interactions with those samples and modalities.

For the surrogate $\hat g$, the $N$ samples are drawn independently from the same selection procedure, and their observed losses equal $L(f_W;S)$. Assume that the function $S\mapsto L(f_W;S)$ belongs to $\mathcal H_K$ and satisfies $\|L(f_W;\cdot)\|_{\mathcal H_K}\leq\rho$. We show the following statement.

\medskip
\begin{proposition}\label{prop_surrogate_sample_complexity}
For any $\lambda\in(0,1]$ and $\delta\in(0,1)$, there exist universal constants $C_0,C_1>0$ such that, if
$N\geq\frac{C_0}{\lambda}\log\left(\frac{C_0}{\lambda\delta}\right)$,
then, with probability at least $1-\delta$ over the sampled selections,
\[
\mathbb E_S\left[\left(\hat g(S)-L(f_W;S)\right)^2\right]
\leq C_1\lambda\rho^2.
\]
\end{proposition}

This result bounds the surrogate's prediction error in terms of the number of sampled selections and the regularization parameter. We defer the proof to Appendix~\ref{app_surrogate_sample_complexity_proof}.

\subsection{Efficient Estimation}

For scalability, fitting the surrogate does not require retraining the model for every selection state. Inspired by the first-order approximation in previous works~\cite{li2024scalable,malladi2023kernel,park2023trak}, we linearize the multimodal model around a checkpoint learned on all samples and modalities using a first-order approximation. Let $W^\star$ denote this checkpoint, and $f_{W}(\mathbf{x}_i;S_{:,i})$ denote the model output for sample $i$ under the modality selection $S_{:,i}$. For the parameter $W$ close to $W^\star$, we use the first-order Taylor approximation
\begin{equation}
f_{W}(\mathbf{x}_i;S_{:,i})
=
f_{W^\star}(\mathbf{x}_i;S_{:,i})
+
\nabla_{W}f_{W^\star}(\mathbf{x}_i;S_{:,i})(W-W^\star)+\epsilon,
\label{eq_first_order_multimodal}
\end{equation} 
where $\nabla_{W}f_{W^\star}(\mathbf{x}_i;S_{:,i})$ is the gradient with respect to $W$ and $\epsilon$ denotes the approximation error. We compute and cache both the model output and gradient for each selection $S$. 
In Appendix~\ref{app_first_order}, we empirically evaluate $\epsilon$ and find it to be small in the estimation. The objective under $S$ can be approximated by
\begin{equation}
\frac{1}{\sum_{j=1}^n \mathbf{v}^{(j)}(S)} \sum_{i=1}^n \mathbf{v}^{(i)}(S) \, \ell\!\left( f_{W^\star}(\mathbf{x}_i;S_{:,i}) + \nabla_{W}f_{W^\star}(\mathbf{x}_i;S_{:,i})(W-W^\star), y_i
\right).
\label{eq_linearized_state_objective}
\end{equation}
Thus, the loss associated with each sampled $S$ can be estimated by optimizing the linearized objective, without retraining the original multimodal model from scratch. We report the computational speedup achieved by this approximation in Appendix~\ref{app_first_order}.

%% file: experiments.tex
\begin{figure}[t]
    \centering
    \includegraphics[width=1.00\linewidth]{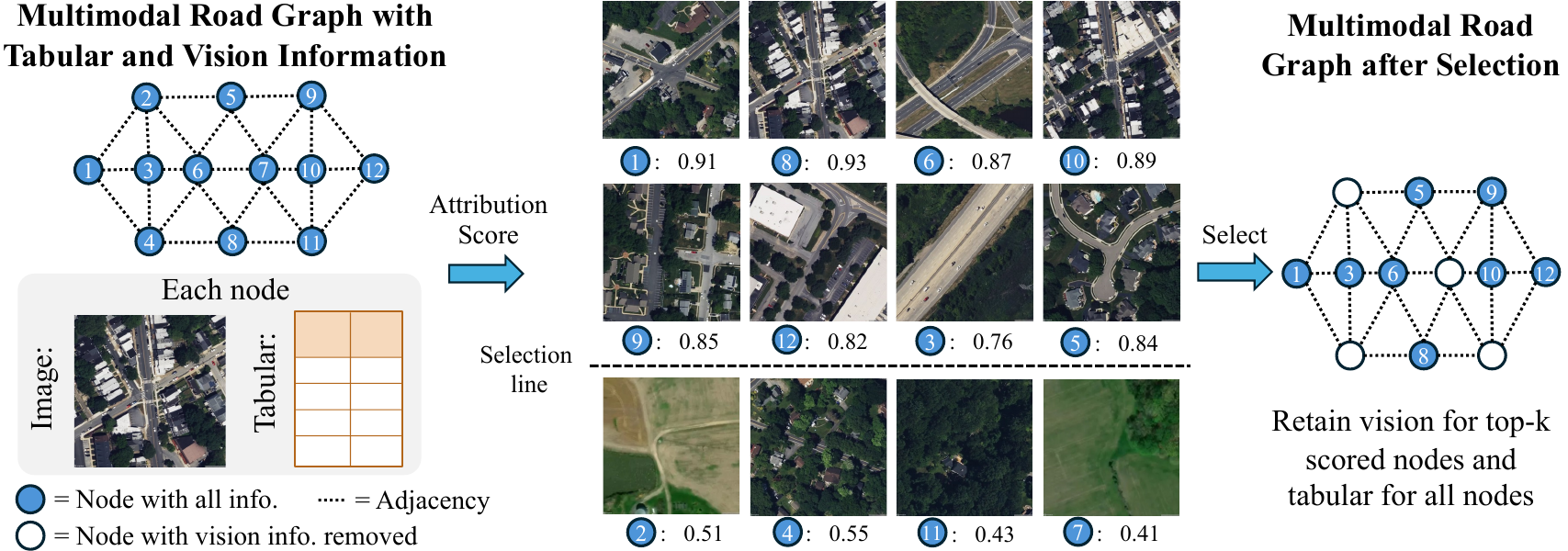}
    \caption{Overview of surrogate-guided data selection on \textsc{MMTraCE}.
    Each node represents a road-network sample with tabular and vision modalities. The structured kernel surrogate assigns an attribution score to each node image. We keep the vision information for the top-$k$ scored nodes.
    }
    \label{fig_kernel_selection}
\end{figure}

\section{Experiments}\label{sec_exp}

Our experiments investigate two key questions. 
First, how does \algo{} compare with existing attribution methods and surrogate-based methods on multimodal attribution benchmarks? 
Second, can the attributions produced by \algo{} be used for downstream data selection? 
We evaluate these questions on MultiBench and \textsc{MMTraCE}. 
The results show that \algo{} achieves the best attribution performance across multiple tasks, and that data selected by \algo{} leads to the highest downstream accuracy under a fixed selection budget. 
We further show through ablation studies that \algo{} is robust to kernel bandwidth choices and generalizes across different modalities and selection rates.

\subsection{Experimental Setup}

\textbf{Datasets.} We first evaluate our framework on the spatiotemporal traffic accident prediction task \textsc{MMTraCE}~\cite{zhang2026learning}. It contains a US state-level road network, accompanied by weather, traffic volume, satellite images, and longitudinal accident records from six states. It covers tabular and vision modalities, and the objective is to predict accident occurrences on a target edge. Then, we evaluate on six real-world tasks in MultiBench~\cite{liang2021multibench}: UR-FUNNY (Humor), MUStARD (Sarcasm), CMU-MOSI (Mosi), CMU-MOSEI (Mosei), AV-MNIST (Avmnist), MM-IMDB (Imdb). These datasets cover diverse modalities, including text, audio, vision, and tabular data.

\textbf{Baselines.} We compare our method with data attribution methods including Influence functions~\cite{koh2017understanding}, TracIn~\cite{pruthi2020estimating}, Trak~\cite{park2023trak}, DSIR~\cite{xie2023data}, LESS~\cite{xia2024less}, and BIF~\cite{kreer2025bayesian}. Lastly, we compare our approach with linear surrogate models~\cite{liidentification}. 

\textbf{Evaluation.} We denote the evaluated sampling states by
$\{S_p\}_{p=1}^N$, with corresponding test losses $y_p$. The predicted test loss from an attribution method is defined as $\hat{g}(S_p)$. We first evaluate the LDS score for all methods. It is defined as the Spearman correlation coefficient between the predicted ranking and the real ranking: $\mathrm{LDS} = \mathrm{Spearman} \left( \{\hat{g}(S_p)\}_{p=1}^N, \{y_p\}_{p=1}^N \right)$. A higher LDS score indicates that the attribution method better recovers the ranking of subset-level test losses.

Then, we frame data selection as a downstream evaluation of attribution methods under an asymmetric setting. The model uses all data from the first modality and only a $20\%$ subset from the second modality. Each attribution method scores samples in the second modality, and we select the top $20\%$ as the retained subset. We then fit the multimodal model with the complete first modality and the selected second-modality subset, and evaluate it on a standard test set. We add an illustrative example in Figure~\ref{fig_kernel_selection} for better understanding.

\subsection{Experimental Results}

\textbf{Attribution results.} 
As shown in Table~\ref{tab_traffic_attribution}, \algo{} achieves the highest LDS score across both \textsc{MMTraCE} and MultiBench. Compared with the strongest baseline, \algo{} improves the per-task relative LDS by $50.1\%$ on \textsc{MMTraCE} and by $65.6\%$ on MultiBench. The gains are especially large on more challenging multimodal tasks, such as Mosi and Sarcasm, where linear attribution is less able to capture nonlinear subset-level effects. These results show that structured nonlinear surrogate modeling is important for recovering the ranking of subset-level test losses in multimodal prediction.

\begin{table}[t]
\centering
\caption{Results on multimodal attribution across \textsc{MMTraCE} and MultiBench. We report LDS scores for six \textsc{MMTraCE} states and six MultiBench tasks. All results are reported as the mean and standard deviation over three independent runs.
}\label{tab_traffic_attribution}
{\small
\begin{tabular}{l|cccccc}
\toprule
Methods              & DE & MA & MD & NV & MT & IA \\
\midrule
Influence functions         & $0.22_{\pm 0.03}$ & $0.12_{\pm 0.05}$ & $0.17_{\pm 0.03}$ & $0.10_{\pm 0.02}$ & $0.23_{\pm 0.05}$ & $0.07_{\pm 0.01}$ \\
TracIn                      & $0.29_{\pm 0.05}$ & $0.11_{\pm 0.08}$ & $0.21_{\pm 0.04}$ & $0.25_{\pm 0.05}$ & $0.16_{\pm 0.07}$ & $0.12_{\pm 0.01}$ \\
Trak                        & $0.26_{\pm 0.01}$ & $0.12_{\pm 0.01}$ & $0.14_{\pm 0.02}$ & $0.19_{\pm 0.03}$ & $0.24_{\pm 0.01}$ & $0.16_{\pm 0.07}$ \\
BIF                        & $0.31_{\pm 0.04}$ & $0.18_{\pm 0.02}$ & $0.24_{\pm 0.02}$ & $0.18_{\pm 0.03}$ & $0.17_{\pm 0.04}$ & $0.21_{\pm 0.03}$ \\
Linear surrogates               & $0.46_{\pm 0.06}$ & $0.51_{\pm 0.02}$ & $0.41_{\pm 0.01}$ & $0.42_{\pm 0.10}$ & $0.24_{\pm 0.06}$ & $0.58_{\pm 0.02}$ \\
\algo{}  (Algorithm \ref{alg_srbf})                   & $\mathbf{0.64}_{\pm 0.03}$ & $\mathbf{0.62}_{\pm 0.05}$ & $\mathbf{0.59}_{\pm 0.10}$ & $\mathbf{0.58}_{\pm 0.02}$ & $\mathbf{0.57}_{\pm 0.06}$ & $\mathbf{0.70}_{\pm 0.03}$ \\
\midrule
Methods              & Avmnist & Humor & Imdb & Mosei & Mosi & Sarcasm \\
\midrule
Influence functions  & $0.14_{\pm 0.14}$ & $0.11_{\pm 0.21}$ & $0.23_{\pm 0.16}$ & $0.27_{\pm 0.25}$ & $0.17_{\pm 0.11}$ & $0.17_{\pm 0.22}$ \\
TracIn               & $0.05_{\pm 0.07}$ & $0.12_{\pm 0.18}$ & $0.20_{\pm 0.18}$ & $0.04_{\pm 0.13}$ & $0.17_{\pm 0.17}$ & $0.12_{\pm 0.10}$ \\
Trak                 & $0.28_{\pm 0.16}$ & $0.21_{\pm 0.14}$ & $0.33_{\pm 0.15}$ & $0.04_{\pm 0.08}$ & $0.07_{\pm 0.20}$ & $0.05_{\pm 0.11}$ \\
BIF                 & $0.30_{\pm 0.12}$ & $0.19_{\pm 0.11}$ & $0.32_{\pm 0.09}$ & $0.14_{\pm 0.07}$ & $0.16_{\pm0.11}$ & $0.20_{\pm 0.13}$ \\
Linear surrogates        & $0.56_{\pm 0.10}$ & $0.47_{\pm 0.05}$ & $0.27_{\pm 0.07}$ & $0.66_{\pm 0.05}$ & $0.21_{\pm 0.17}$ & $0.30_{\pm 0.24}$ \\
\algo{}  (Algorithm \ref{alg_srbf})             & $\mathbf{0.73}_{\pm 0.04}$ & $\mathbf{0.69}_{\pm 0.06}$ & $\mathbf{0.43}_{\pm 0.05}$ & $\mathbf{0.73}_{\pm 0.03}$ & $\mathbf{0.62}_{\pm 0.01}$ & $\mathbf{0.54}_{\pm 0.06}$ \\
\bottomrule
\end{tabular}}
\end{table}

\textbf{Data selection results.}
Given the strong attribution performance, we further evaluate \algo{} on downstream data selection. Under the fixed top-$20\%$ vision-modality selection budget, \algo{} achieves the best test accuracy on both \textsc{MMTraCE} and MultiBench, as shown in Table~\ref{tab_traffic_data_selection}. Compared with the strongest baseline for each state, \algo{} improves the average accuracy by $1.52\%$ on \textsc{MMTraCE}. On MultiBench, \algo{} also improves over the strongest baseline for each task by $2.33\%$ on average, with the largest gain on AV-MNIST. These results show that the learned sample-modality attributions can effectively guide downstream data selection across both traffic prediction and general multimodal benchmarks.

\subsection{Ablation Studies}
\label{sec_ablation}
\textbf{Ablation for the data attribution.} 
We ablate the surrogate modeling methods in \algo{}. We test the performance of the flat surrogate kernel instead of the structured kernel. As shown in Figure~\ref{fig_ablation_surrogate}, \algo{} performs better than the flat surrogate kernel on DE, MA, and MD, with a $9.6\%$ LDS improvement. Similar results are found on MultiBench, and we defer the results to Figure~\ref{fig_surrogate_omitted}.
We then ablate the sample-level bandwidth $\gamma_1$ and the modality-level bandwidth $\gamma_2$ of the structured surrogate on \textsc{MMTraCE}. Across a broad range of bandwidth values, the LDS score remains stable. 

\begin{table}[t]
\centering
\caption{Test accuracy of data selection methods on \textsc{MMTraCE} \cite{zhang2026learning} and MultiBench \cite{liang2021multibench}. Each approach selects the top $20\%$ vision-modality samples under a fixed selection budget, and we report the mean and standard deviation over three independent runs.
}\label{tab_traffic_data_selection}
{\small
\begin{tabular}{l|cccccc}
\toprule
Methods              & DE & MA & MD & NV & MT & IA \\
\midrule
Influence functions         & $71.0_{\pm 0.1}$ & $77.3_{\pm 0.8}$ & $78.1_{\pm 0.3}$ & $85.9_{\pm 0.1}$ & $78.7_{\pm 0.2}$ & $75.4_{\pm 0.1}$ \\
DSIR                        & $70.2_{\pm 0.8}$ & $76.1_{\pm 0.5}$ & $77.9_{\pm 0.7}$ & $85.2_{\pm 0.8}$ & $77.5_{\pm 0.7}$ & $75.7_{\pm 0.4}$ \\
TracIn                      & $72.1_{\pm 0.4}$ & $77.9_{\pm 0.3}$ & $79.5_{\pm 0.4}$ & $85.4_{\pm 0.1}$ & $79.5_{\pm 0.1}$ & $76.4_{\pm 0.2}$ \\
Trak                        & $71.3_{\pm 0.5}$ & $78.2_{\pm 0.2}$ & $79.5_{\pm 0.2}$ & $86.1_{\pm 0.0}$ & $78.5_{\pm 0.2}$ & $76.5_{\pm 0.1}$ \\
BIF                         & $71.4_{\pm 0.3}$ & $78.0_{\pm 0.3}$ & $79.7_{\pm 0.3}$ & $85.8_{\pm 0.2}$ & $78.9_{\pm 0.3}$ & $76.6_{\pm 0.2}$ \\
LESS                        & $71.4_{\pm 0.0}$ & $78.0_{\pm 0.3}$ & $79.6_{\pm 0.2}$ & $86.4_{\pm 0.2}$ & $79.5_{\pm 1.2}$ & $76.9_{\pm 0.8}$ \\
Linear surrogate models               & $71.8_{\pm 0.4}$ & $78.2_{\pm 0.1}$ & $79.9_{\pm 0.1}$ & $86.1_{\pm 0.1}$ & $79.1_{\pm 0.3}$ & $76.6_{\pm 0.1}$ \\
\algo{} (Algorithm \ref{alg_srbf})                    & $\mathbf{73.6}_{\pm 0.3}$ & $\mathbf{79.6}_{\pm 0.1}$ & $\mathbf{81.3}_{\pm 0.1}$ & $\mathbf{88.1}_{\pm 0.2}$ & $\mathbf{81.6}_{\pm 0.2}$ & $\mathbf{77.9}_{\pm 0.3}$ \\
\midrule
Methods              & Avmnist & Humor & Imdb & Mosei & Mosi & Sarcasm \\
\midrule
Influence functions     & $34.9_{\pm 0.9}$ & $56.6_{\pm 1.1}$ & $52.8_{\pm 1.1}$ & $62.9_{\pm 1.2}$ & $67.8_{\pm 1.0}$ & $56.5_{\pm 1.9}$ \\
DSIR                    & $39.7_{\pm 0.4}$ & $53.1_{\pm 0.7}$ & $51.9_{\pm 0.8}$ & $60.9_{\pm 0.9}$ & $62.3_{\pm 1.1}$ & $56.8_{\pm 1.5}$ \\
TracIn                  & $52.9_{\pm 0.6}$ & $56.6_{\pm 1.1}$ & $51.9_{\pm 1.4}$ & $63.2_{\pm 0.8}$ & $60.9_{\pm 1.5}$ & $53.6_{\pm 2.1}$ \\
Trak                    & $53.9_{\pm 0.8}$ & $56.4_{\pm 0.9}$ & $52.8_{\pm 0.9}$ & $60.6_{\pm 1.1}$ & $66.2_{\pm 0.8}$ & $58.0_{\pm 1.5}$ \\
BIF                 & $53.5_{\pm 1.0}$ & $56.7_{\pm 0.7}$ & $52.5_{\pm 1.1}$ & $61.7_{\pm 0.9}$ & $67.5_{\pm 0.9}$ & $57.8_{\pm 1.8}$ \\
LESS                    & $53.3_{\pm 0.6}$ & $56.4_{\pm 0.8}$ & $52.8_{\pm 1.0}$ & $60.8_{\pm 0.9}$ & $66.3_{\pm 1.2}$ & $57.6_{\pm 1.2}$ \\
Linear surrogate models       & $56.5_{\pm 1.2}$ & $56.6_{\pm 0.8}$ & $52.9_{\pm 1.0}$ & $63.3_{\pm 0.8}$ & $68.0_{\pm 0.9}$ & $57.2_{\pm 1.0}$ \\
\algo{} (Algorithm \ref{alg_srbf})                 & $\mathbf{61.5}_{\pm 0.9}$ & $\mathbf{58.5}_{\pm 0.5}$ & $\mathbf{53.9}_{\pm 1.3}$ & $\mathbf{65.4}_{\pm 1.4}$ & $\mathbf{69.2}_{\pm 1.1}$ & $\mathbf{60.9}_{\pm 1.5}$ \\
\bottomrule
\end{tabular}}
\end{table}

\textbf{Ablation analysis on data selection.}
For data selection, we conduct two ablation studies on the \textsc{MMTraCE} dataset.
First, we change the selected modality from vision to tabular information while keeping the same top-$20\%$ selection budget. 
As shown in Figure~\ref{fig_ablation_modality}, the accuracy drops noticeably compared with vision selection in Table~\ref{tab_traffic_data_selection}. This indicates that tabular information is more critical for traffic accident prediction than satellite vision features. 
\algo{} still performs the best on DE, MA, and MD, with a $3.4\%$ accuracy improvement. This shows that the learned attribution remains effective across different modalities. We defer the ablation results for the rest of the states to Figure~\ref{fig_modality_omitted}.

Second, we vary the selection rate for the selected modality. 
Figure~\ref{fig_ablation_sample_rate} shows that accuracy improves as the selection rate increases, but the gain gradually saturates at larger budgets. This suggests that the most useful samples are selected early, and that additional samples yield diminishing returns.

\subsection{Local Geometric Measures}

Next, we evaluate the cross-state generalization results of applying \algo{} on \textsc{MMTraCE}. Specifically, we learn the model on one source state using the downsampled data selected by \algo{} and test it on each target state. As shown in Figure~\ref{fig_cross_state_heatmap}, the selected subsets generalize well across different source-target pairs. Among the source states, models based on MT data achieve stronger average performance than those based on DE or MA data, with an average accuracy of $74.9\%$. This shows that the selected MT subset may contain more transferable information for cross-state accident prediction.

Then, we assess model generalization through loss sharpness. A flatter loss landscape suggests that the selected subset induces a model less sensitive to small input perturbations and more likely to generalize across states. Since directly computing the Hessian trace is expensive, we estimate it by adding small isotropic Gaussian noise and measuring the expected loss increase.

We use the trace of the Hessian matrix of the model for evaluating the generalization ability~\cite{ju2022robust,zhang2024noise}. We compare the model from different states in MMTraCE with the data selected by different attribution methods. As shown in Table~\ref{tab_hessian_trace}, \algo{} not only achieves the lowest test loss, but also has the lowest Hessian trace value. This shows that the model on data selected by \algo{} generalizes better than models on data selected by other attribution methods.

\begin{table}[h]
\centering
\caption{The Hessian trace and test loss of the model based on Delaware, Massachusetts, Maryland, and Nevada. We compare \algo{} against several attribution methods including influence functions \cite{koh2017understanding}, TracIn~\cite{pruthi2020estimating}, Trak \cite{park2023trak} and BIF~\cite{kreer2025bayesian}.}\label{tab_hessian_trace}
\resizebox{0.99\columnwidth}{!}
{
\begin{tabular}{l|cc|cc|cc|cc}
\toprule
Methods              & \multicolumn{2}{c|}{DE} & \multicolumn{2}{c|}{MA} & \multicolumn{2}{c|}{MD} & \multicolumn{2}{c}{NV}\\
Metric & Trace & Loss & Trace & Loss & Trace & Loss & Trace & Loss \\
\midrule
Influence functions & $3.88\times10^{2}$ & $0.61$ & $8.75\times10^{2}$ & $0.84$ & $6.17\times10^{3}$ & $1.04$ & $2.53\times10^{2}$ & $0.70$ \\
TracIn & $4.69\times10^{2}$ & $0.62$ & $3.63\times10^{2}$ & $0.90$ & $1.19\times10^{4}$ & $1.14$ & $2.74\times10^{2}$ & $0.74$ \\
Trak & $5.45\times10^{2}$ & $0.60$ & $3.27\times10^{2}$ & $0.87$ & $1.90\times10^{3}$ & $1.11$ & $3.95\times10^{2}$ & $0.71$ \\
BIF & $3.76\times10^{2}$ & $0.61$ & $1.47\times10^{3}$ & $0.81$ & $7.84\times10^{3}$ & $1.11$ & $4.70\times10^{2}$ & $0.74$ \\
\algo{} (Algorithm \ref{alg_srbf}) & $\bf{2.46\times10^{2}}$ & $\bf{0.59}$ & $\bf{2.68\times10^{2}}$ & $\bf{0.79}$ & $\bf{1.59\times10^{3}}$ & $\bf{0.98}$ & $\bf{7.35\times10^{1}}$ & $\bf{0.67}$ \\
\bottomrule
\end{tabular}}
\end{table}

%% file: related.tex
\begin{figure}[t]
\centering
\begin{subfigure}{0.24\linewidth}
    \centering
    \includegraphics[width=\linewidth]{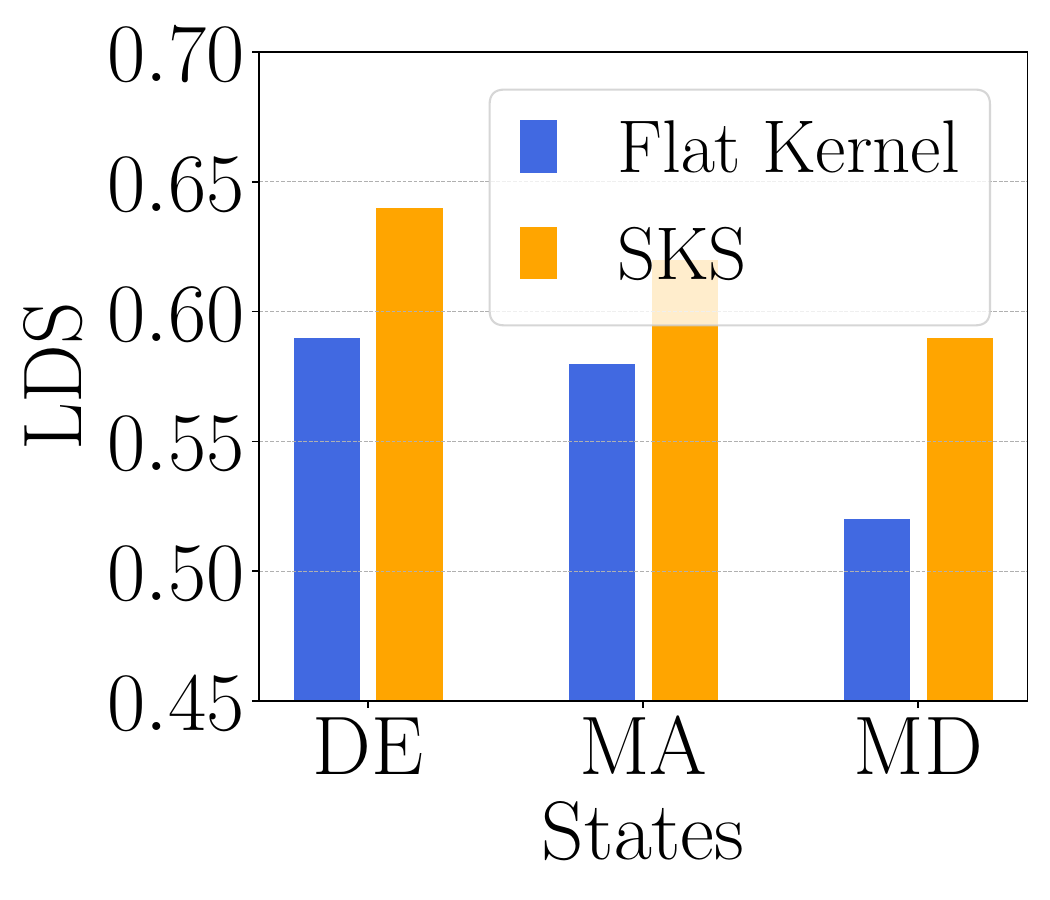}
    \caption{Compositional kernel}
    \label{fig_ablation_surrogate}
\end{subfigure}\hfill
\begin{subfigure}{0.24\linewidth}
    \centering
    \includegraphics[width=\linewidth]{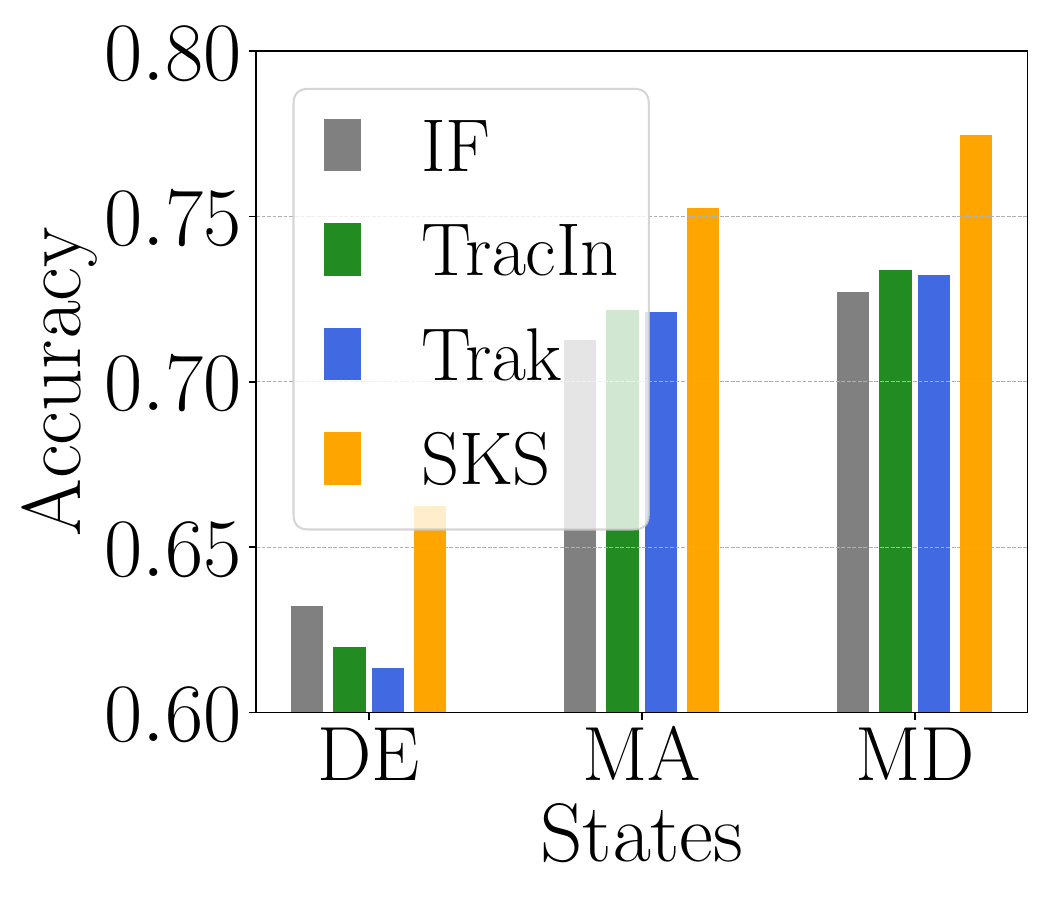}
    \caption{Data modality}
    \label{fig_ablation_modality}
\end{subfigure}\hfill
\begin{subfigure}{0.24\linewidth}
    \centering
    \includegraphics[width=\linewidth]{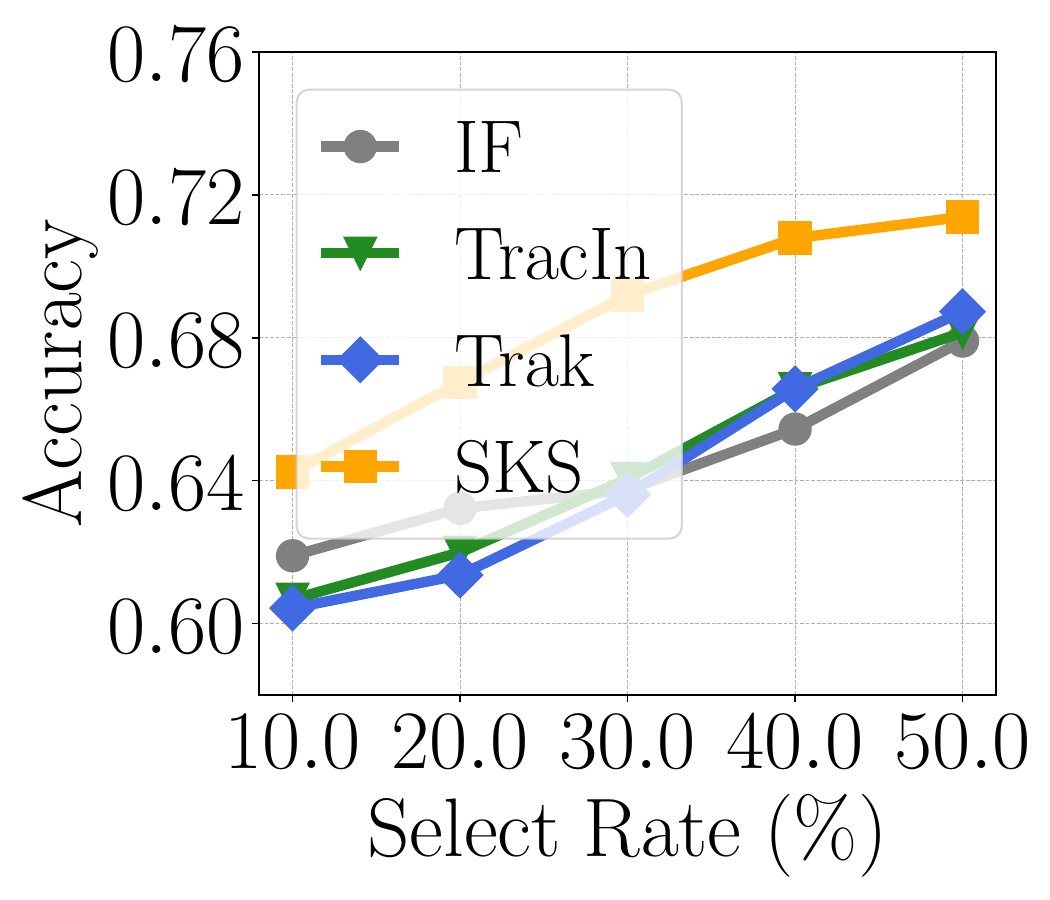}
    \caption{Selection rate}
    \label{fig_ablation_sample_rate}
\end{subfigure}\hfill
\begin{subfigure}{0.27\linewidth}
    \centering
    \includegraphics[width=0.8\linewidth]{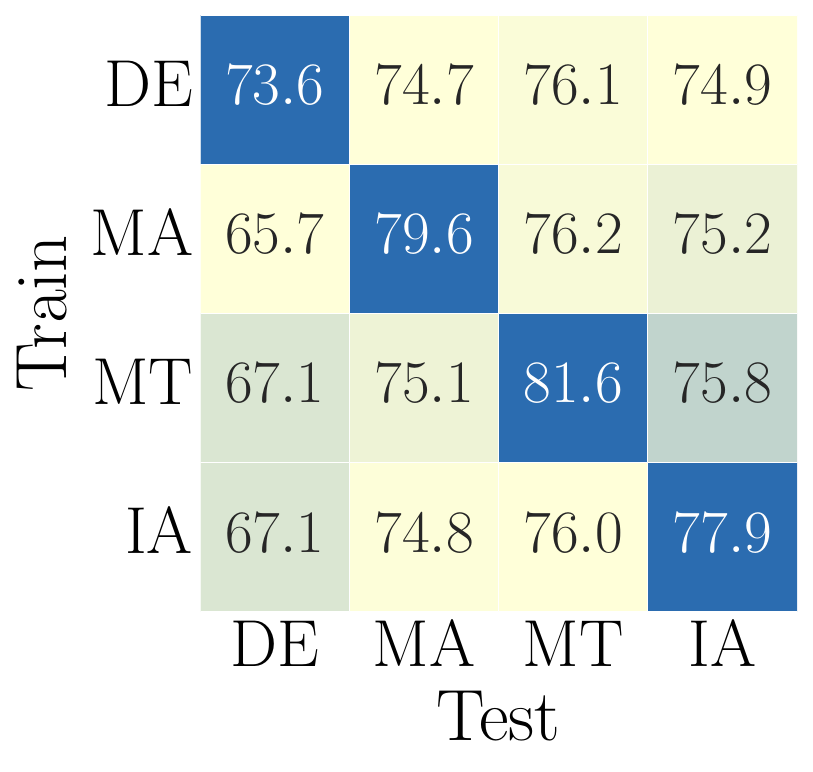}
    \caption{Cross-state generalization}
    \label{fig_cross_state_heatmap}
\end{subfigure}
\caption{Ablation studies on the \textsc{MMTraCE} dataset \cite{zhang2026learning}. Figure~\ref{fig_ablation_surrogate} evaluates surrogate modeling by comparing the LDS of \algo{} with the flat kernel on Delaware, Massachusetts, and Maryland. Figure~\ref{fig_ablation_modality} studies the effect of data modalities. We retain all visual data, select $20\%$ of the tabular data using each method, and report the resulting test accuracy. Figure~\ref{fig_ablation_sample_rate} evaluates selection rates ranging from $10\%$ to $50\%$ and reports the test accuracy on Delaware. Figure~\ref{fig_cross_state_heatmap} evaluates cross-state generalization by fitting the model on one state and directly testing it on another state.
}\label{fig_ablation}
\end{figure}

\section{Related Work}

\textbf{Multimodal representation learning.} 
Multimodal learning relies on relationships between information sources to improve prediction. Different modalities can compensate for missing or unreliable information \cite{yang1999multimodal} and provide context needed to interpret one another \cite{yang2001automatic,zhao2026papermind}. Learning these relationships provides a basis for constructing useful representations and reducing the number of samples required \cite{kakade2007multi}.
\citet{bei2026mem} show that answering questions about past conversations requires connecting visual details with facts introduced in other dialogue turns. It is important for multi-step search~\cite{ning2026mcsearch}, since evidence from one modality can guide what to retrieve from another.
The reliability of these relationships also matters. \citet{sun2024aligning} model uncertainty in image-text alignment to capture whether the available information supports a connection between their meanings.

These relationships guide representation learning by identifying which information should be preserved and how it should be organized. For example, semantic correspondence between language and vision guides representations to emphasize visual content relevant to the text and bring their meanings into alignment \cite{chen2026cvp,khan2023contrastive}. To connect representations to prediction, deep supervision makes representations themselves predictive rather than relying solely on the final output \cite{lee2015deeply}, and knowledge transfer through representations supplies localization information missing from target labels \cite{wang2023d}.

Missing information and temporal changes also shape what representations need to capture. The absence of one modality can itself carry information, so representations should account for what is missing as well as what is observed \cite{liang2025causal}. For evolving states, this missingness information helps refine estimates based on past observations, so that representations can track changes at each time step \cite{liang2026learning}. Beyond using observations to track changing states, representations can model how feedback drives subsequent changes \cite{pan2026two}.

\paragraph{Algorithms for influence estimation.}
The problem of influence estimation aims to quantify how individual examples affect predictions. Classical influence functions~\citep{koh2017understanding} estimate the effect of upweighting or removing one sample through a local second-order approximation. However, estimating inverse Hessian-vector products is still expensive for large models. To reduce computation, gradient similarity across  checkpoints avoids explicit Hessian computation \cite{pruthi2020estimating}, and model linearization with random projections enables attribution at larger scales \cite{park2023trak}. Other than computational efficiency, attribution also raises the question of whether estimated importance rankings are reliable. For feature attribution, \citet{goldwasser2024statistical} address this question by testing whether differences in estimated scores support the reported rankings despite sampling uncertainty.

This type of influence analysis also extends naturally from quantifying individual examples to entire evaluation tasks. 
In multi-task learning, joint training can improve or hurt a task's performance depending on the tasks' input distributions and sample sizes~\cite{wu2020understanding,yang2025precise}. \citet{liidentification} use surrogate models to estimate the effects of different task combinations on performance, identifying 
which combinations lead to positive or negative transfer. Then, \citet{li2023boosting,li2024scalablemtl} use estimated transfer relationships to identify task groups that benefit from joint training.
The cost of comparing combinations motivates first-order approximations of fine-tuning performance \cite{li2025efficient,zhang2025linear}. These estimates inform both which tasks to group during optimization \cite{zhang2025scalable,li2026moplex} and where they should share parameters within a model \cite{li2025efficiently,zhang2026long}. Theoretical analysis in the setting of high-dimensional linear regression shows when learning a shared linear model on source and target data improves target-task prediction compared with fitting the model on target data alone 
\cite{yang2025precise}.

\paragraph{Information-theoretic perspectives on multimodal learning.}
Information-theoretic analyses relate the benefits of multimodal learning to how predictive information is distributed across modalities. Shared information allows unlabeled paired data to constrain learning \cite{sridharan2008information}, and bounds on information available only through joint observation quantify the potential gains from combining modalities \cite{liangmultimodal}. Beyond information content, analyses of data augmentation distinguish gains from expanded data coverage and regularization \cite{wu2020generalization}. Complementary optimization work addresses the tradeoff between data reuse and the distribution mismatch induced by repeated updates \cite{wang2025slow}. These perspectives characterize conditions for effective data use, but do not directly quantify how individual samples and their modalities affect test performance.

\section{Conclusion}

We study cross-sample modality attribution in multimodal learning. We propose a new algorithm that estimates the subset-to-loss response while preserving the hierarchy between sample participation and within-sample modality composition. We further apply the learned sample-modality attributions to downstream data selection. Across \textsc{MMTraCE} and MultiBench, our approach consistently improves over prior work in terms of attribution quality and data selection accuracy. Further geometric analysis shows that models trained on data selected by our proposed algorithm have a lower Hessian trace, suggesting a flatter loss surface and better generalization.

\section*{Acknowledgment}

This work received financial support from NSF award IIS-2412008 and a startup fund from Northeastern University.
Ziniu Zhang and Zhenshuo Zhang received additional financial support from Northeastern University's Khoury PhD fellowships.
R. Xiong is supported in part by the URC award at Emory University.

%% file: appendix.tex
\section{Omitted Materials from Section \ref{sec_main}}\label{app_proof}

We first introduce the notations.
Let $\mathcal D=\{(\mathbf{x}_i,y_i)\}_{i=1}^n$ denote a training dataset of $n$ samples, where $\mathbf{x}_i=(\mathbf{x}_i^{(1)},\ldots,\mathbf{x}_i^{(M)})$ contains $M$ input modalities and $y_i$ is the corresponding output. The selection state matrix $S$ specifies which modalities are retained in each sample, and $\mathbf v(S)$ indicates which samples retain at least one modality under $S$. We write $f_W$ for the multimodal model with parameters $W$, $\ell$ for the per-sample loss, and $L(f_W;S)$ for the test loss of the model trained under $S$. 

The RBF kernels $k_1$ and $k_2$, with bandwidths $\gamma_1$ and $\gamma_2$, measure similarity in sample participation and within-sample modality selection, respectively. Combining these kernels gives the structured kernel $K(S,S')$ for two different selections. Let $\mathcal H_K$ be the reproducing kernel Hilbert space induced by $K$, we fit a surrogate function $g\in\mathcal H_K$ using $N$ sampled combinations of retained samples and modalities. In kernel ridge regression, $\mathbf K$ is the Gram matrix with entries $K_{pq}=K(S_p,S_q)$, $\mathbf k_S$ contains the kernel similarity between a query selection state $S$ and the $N$ sampled states, and $\lambda$ controls the ridge regularization.

For a selection state $S$, $A^{(j,m)}(S)$ denotes the attribution of modality $m$ in sample $j$. These attribution values are used to rank and select sample-modality pairs for downstream model training.

Finally, we use $B_M$ as the Gram matrix over its nonzero $M$-modality states. The constant $\rho$ bounds the RKHS norm of the mapping $S\mapsto L(f_W;S)$, so that $\|L(f_W;\cdot)\|_{\mathcal H_K}\leq\rho$. We use $\epsilon$ to denote the error introduced by the first-order approximation.

\subsection{Proof of Proposition~\ref{prop_surrogate_sample_complexity}}
\label{app_surrogate_sample_complexity_proof}

\begin{proof}
Since $\hat g$ minimizes the kernel ridge regression objective over
$\mathcal H_K$, its objective value cannot exceed that obtained by
choosing $g=L(f_W;\cdot)\in\mathcal H_K$. Thus,
\[
\begin{aligned}
&\frac{1}{N}\sum_{p=1}^N \left(\hat g(S_p)-L(f_W;S_p)\right)^2 +\lambda\|\hat g\|_{\mathcal H_K}^2 \\
&\leq \frac{1}{N}\sum_{p=1}^N \left(L(f_W;S_p)-L(f_W;S_p)\right)^2 +\lambda\|L(f_W;\cdot)\|_{\mathcal H_K}^2 \\
&= \lambda\|L(f_W;\cdot)\|_{\mathcal H_K}^2 \leq \lambda\rho^2,
\end{aligned}
\]
where the last inequality uses the assumption
$\|L(f_W;\cdot)\|_{\mathcal H_K}\leq\rho$.
Thus,
\[
\frac{1}{N}\sum_{p=1}^N
\left(\hat g(S_p)-L(f_W;S_p)\right)^2
\leq\lambda\rho^2,
\qquad
\|\hat g(\cdot)\|_{\mathcal H_K}\leq\rho.
\]
By the triangle inequality, together with $\|\hat g(\cdot)\|_{\mathcal H_K}\leq\rho$ and the assumption $\|L(f_W;\cdot)\|_{\mathcal H_K}\leq\rho$, we obtain
\[
\|\hat g(\cdot)-L(f_W;\cdot)\|_{\mathcal H_K}
\leq \|\hat g(\cdot)\|_{\mathcal H_K}
+\|L(f_W;\cdot)\|_{\mathcal H_K}
\leq \rho+\rho
=2\rho.
\]

Since $S_1,\ldots,S_N$ are independent, $K(S,S)=1$, and
\[
N\geq\frac{C_0}{\lambda}
\log\left(\frac{C_0}{\lambda\delta}\right)
\]
for a sufficiently large universal constant $C_0$, the operator Bernstein inequality bounds the expected squared error by the empirical squared error and a regularization term. Specifically, with probability at least $1-\delta$, simultaneously for every $g\in\mathcal H_K$,
\[
\mathbb E_S\left[\left(g(S)-L(f_W;S)\right)^2\right] \leq \frac{2}{N}\sum_{p=1}^N \left(g(S_p)-L(f_W;S_p)\right)^2 +\lambda\|g(\cdot)-L(f_W;\cdot)\|_{\mathcal H_K}^2.
\]
Applying this inequality to $\hat g(\cdot)$ and using the bounds above yields
\[
\mathbb E_S\left[
\left(\hat g(S)-L(f_W;S)\right)^2
\right]
\leq 2\lambda\rho^2+4\lambda\rho^2
=6\lambda\rho^2.
\]
This proves the claim with $C_1=6$.
\end{proof}

\subsection{Positive Semidefiniteness of the Structured Kernel}\label{app_psd}

We use $K$ to measure similarity between different selections and fit a loss predictor through kernel ridge regression. To establish its validity as a kernel, for a sample $j$, let $B_M \in \mathbb R^{(2^M-1)\times(2^M-1)}$ be the Gram matrix indexed by the nonzero modality states $S_{:,j}, S'_{:,j} \in \{0,1\}^M \setminus \{\mathbf 0\}$, with entries
\[ (B_M)_{S_{:,j},S'_{:,j}} = \exp\left(-\gamma_2\|S_{:,j}-S'_{:,j}\|_2^2\right) \]
Let $\mathbf 1 \in \mathbb R^{2^M-1}$ denote the all-ones vector.

\begin{proposition}
\label{prop_psd}
The structured design $K$ is positive semidefinite over all sample and modality selections if
\[ \gamma_2 > 0 \quad\text{and}\quad \gamma_1 \geq \frac{1}{2} \log\!\left(\mathbf 1^\top B_M^{-1}\mathbf 1\right). \]
\end{proposition}

\begin{proof}
To analyze the positive semidefiniteness of $K$, we first rewrite $K$ into an equivalent product form over the samples. Since the first term is
\[\exp \left(-\gamma_1 \|\mathbf{v}(S) - \mathbf{v}(S^{\prime})\|_2^2 \right)=\prod_j\exp \left(-\gamma_1 \|\mathbf{v}^{(j)}(S) - \mathbf{v}^{(j)}(S^{\prime})\|_2^2 \right),\] 
and the second term is
\[
\prod_{j\in[n]:\mathbf{v}^{(j)}(S)=\mathbf{v}^{(j)}(S')=1} \exp \left(-\gamma_2 \|S_{:,j} - S'_{:,j}\|_2^2\right) = \prod_j \exp(- \gamma_2  \mathbf{v}^{(j)}(S)\mathbf{v}^{(j)}(S')  \|S_{:,j} - S_{:,j}'\|_2^2),
\] 
we have the alternative formulation of $K$
\begin{align*}
K(S, S') = \prod_{j=1}^n \exp \left( -\gamma_1 \|\mathbf{v}^{(j)}(S)-\mathbf{v}^{(j)}(S')\|_2^2 - \gamma_2  \mathbf{v}^{(j)}(S)\mathbf{v}^{(j)}(S')  \|S_{:,j} - S_{:,j}'\|_2^2 \right).
\end{align*}
Since $K(S,S')$ is a product over samples, $K$ is positive semidefinite if the factor corresponding to each sample is positive semidefinite. Fix a sample $j$. The $j$-th factor in the product measures the similarity between the modality selections $S_{:,j}$ and $S'_{:,j}$ for this sample. Ordering the possible columns with the zero vector first, followed by all nonzero binary vectors, gives the Gram matrix
\[
\begin{pmatrix}
1 & e^{-\gamma_1}\mathbf 1^\top\\
e^{-\gamma_1}\mathbf 1 & B_M
\end{pmatrix}.
\]
where $\mathbf 1\in\mathbb R^{2^M-1}$ is the all-ones vector and $B_M$ is indexed by the nonzero columns $S_{:,j},S'_{:,j}\in\{0,1\}^M$, with entries
\[
(B_M)_{S_{:,j},S'_{:,j}} = \exp\left(-\gamma_2\|S_{:,j}-S'_{:,j}\|_2^2\right).
\]
Since $B_M$ is the Gram matrix of an RBF kernel on distinct points, it is positive definite for $\gamma_2>0$. By the Schur complement, the block matrix above is positive semidefinite if and only if
\[
1-e^{-2\gamma_1}\mathbf 1^\top B_M^{-1}\mathbf 1\geq 0,
\]
which is equivalent to
\[
\gamma_1\geq\frac{1}{2}\log\left(\mathbf 1^\top B_M^{-1}\mathbf 1\right).
\]
Under this condition, the factor corresponding to sample $j$ is positive semidefinite over all possible columns $S_{:,j}$. Since $K$ is a product of these factors, it is also positive semidefinite.

\end{proof}

Thus, kernel ridge regression has a unique solution for any finite collection of sampled selections and any positive regularization parameter.

\subsection{Extensions}\label{app_extension}

Our structured kernel surrogate naturally extends from two modalities to $M>2$ modalities by defining the global sample-modality selection state as
\[
S =
\begin{bmatrix}
S_{1,1} & \cdots & S_{1,n}\\
\vdots & \ddots & \vdots\\
S_{M,1} & \cdots & S_{M,n}
\end{bmatrix}
\in \{0,1\}^{M\times n},
\]
where $S_{m,j}=1$ if modality $m$ of sample $j$ is retained, and $S_{m,j}=0$ otherwise. The column $S_{:,j}\in\{0,1\}^{M}$ specifies the modalities retained in sample $j$, and $\mathbf v^{(j)}(S)= \max(S_{:,j})$ indicates whether this sample retains at least one modality.

The structured kernel retains the same form as in the two-modality case. The sample-level kernel measures the distance between $\mathbf{v}(S)$ and $\mathbf{v}(S')$, and the modality-level kernel measures the distance between the corresponding $M$-dimensional states $S_{:,j}$ and $S'_{:,j}$. The positive semidefiniteness condition changes only through the local state space. Specifically, $B_M$ now covers the $2^M-1$ nonzero modality states in $\{0,1\}^M$. The structured kernel is positive semidefinite if
$ \gamma_2 > 0$,  and  $\gamma_1 \geq \frac{1}{2} \log\!\left(\mathbf 1^\top B_M^{-1}\mathbf 1\right).$
Under this condition, \algo{} extends directly to $M$ modalities using the same kernel ridge regression and attribution procedures.

\section{Omitted Experimental Details and Results}\label{app_exp}

\paragraph{Models and hyperparameters.} For MultiBench, we use the cross-attention network to fuse the pre-computed modality embeddings provided by the datasets. The hyperparameters are listed in Table~\ref{tab_hyper_multibench}.
For \textsc{MMTraCE}, we use a three-layer GNN encoder and an MLP predictor. The vision embedding is fetched by pretrained CLIP and ViT models. We report the hyperparameters in Table~\ref{tab_hyper_mmtrace}. 

The models we use for validating the first-order approximation are
\href{https://huggingface.co/Qwen/Qwen2-VL-2B}{Qwen2-VL-2B},
\href{https://huggingface.co/Qwen/Qwen3-VL-8B-Instruct}{Qwen3-VL-8B},
\href{https://huggingface.co/Qwen/Qwen3-VL-30B-A3B-Instruct}{Qwen3-VL-30B-A3B},
and \href{https://huggingface.co/meta-llama/Llama-3.2-11B-Vision}{Llama-3.2-11B-Vision}.
All the experiments are conducted on one NVIDIA RTX A6000 GPU and one NVIDIA B200 GPU. For both \textsc{MMTraCE} and MultiBench, we fit the surrogate model using $N=100$ pairs of selections and crossponding losses $\{S_p,L(f_W;S_p)\}_{p=1}^N$.

\begin{table}[ht]
\centering
\caption{Hyperparameters of the Cross-Attention model.}\label{tab_hyper_multibench}
\begin{tabular}{lc}
\toprule
\textbf{Hyperparameter} & \textbf{Value} \\
\hline
Projection dimension & $256$ \\
Number of attention heads  & $4$ \\
Number of layers & $2$ \\
Batch size & $4096$ \\
Dropout probability & $0.1$ \\
Number of epochs & $15$ \\
Learning rate & $5 \times 10^{-4}$ \\
Optimizer & AdamW \\
Weight decay & $1 \times 10^{-4}$ \\
\bottomrule
\end{tabular}
\end{table}

\begin{table}[ht]
\centering
\caption{Hyperparameters of the GNN model used in \textsc{MMTraCE}.}\label{tab_hyper_mmtrace}
\begin{tabular}{lc}
\toprule
\textbf{Hyperparameter} & \textbf{Value} \\
\hline
Number of GNN layers & $3$ \\
Number of Classifier layers & $2$ \\
Batch size & $4096$ \\
Hidden dimension & $256$ \\
Dropout probability & $0.1$ \\
Number of epochs & $15$ \\
Learning rate & $5 \times 10^{-4}$ \\
Optimizer & AdamW \\
Weight decay & $1 \times 10^{-4}$ \\
\bottomrule
\end{tabular}
\end{table}

Since instance-level selections introduce high variance and computational overhead, we design the surrogate sampling at the group level. We partition the training data into $25$ clusters using K-means. To fit the surrogate model, we generate selections by masking each cluster with probability $0.8$ and compute the loss for each selection. We use 100 selections for each of MMTraCE and MultiBench.

\paragraph{Baselines.} For all baselines, we start with samples containing all available modalities. To evaluate a particular modality in a sample, we replace its input with zeros, keep the remaining modalities of the sample unchanged, and then apply the corresponding baseline method.

Influence functions~\cite{koh2017understanding} estimate how upweighting one masked sample affects the loss of another masked sample. We flatten the modalities across samples into individual training items, treating modality $m$ of sample $i$ ($\mathbf x_i^{(m)}$) as a separate item with label $y_i$. We compute the loss gradient for each item and the gradient of the average test loss. Their negative inner product, weighted by the inverse Hessian of the training objective, gives the influence score for that item.

DSIR~\cite{xie2023data} estimates an importance weight for $\mathbf{x}_i^{(m)}$ by taking the ratio of its likelihood under the target distribution to its likelihood under the training distribution in a reduced feature space. A larger ratio indicates that $\mathbf{x}_i^{(m)}$ is more representative of the target distribution. We use this weight to rank and select sample-modality pairs under the fixed budget.

TracIn~\cite{pruthi2020estimating} estimates pairwise influence by accumulating gradient similarity over multiple training checkpoints. Treating each $\mathbf{x}_i^{(m)}$ as an individual training item with label $y_i$, we compute its loss gradient and the gradient of the average test loss at each saved checkpoint. We then take the inner product between these gradients, weight it by the corresponding learning rate, and sum across checkpoints to obtain the attribution score for $\mathbf{x}_i^{(m)}$. A positive score indicates that training on this item is estimated to reduce the average test loss.

TRAK~\cite{park2023trak} estimates how each training item contributes to a target prediction. Treating each $\mathbf{x}_i^{(m)}$ as an individual training item with label $y_i$, we compute gradients of the log-odds of the correct-class probability at $W^\star$ for both training items and target examples. We randomly project these gradients into a lower-dimensional space and compute their similarity using the inverse Gram matrix of the projected training gradients. Each training item's score is then weighted by one minus the probability assigned to $y_i$. Averaging over target examples gives the attribution score for $\mathbf{x}_i^{(m)}$.

BIF~\cite{kreer2025bayesian} estimates influence through loss covariance over a localized parameter distribution, avoiding the inverse-Hessian computation used by classical influence functions. Starting from $W^\star$, we use stochastic gradient Langevin dynamics to sample model parameters from a local posterior defined by the training loss and a Gaussian prior centered at $W^\star$. At each sampled parameter value $W$, we evaluate $\ell(f_W(\mathbf{x}_i^{(m)}),y_i)$ for each individual training item and the average test loss. We compute their negative empirical covariance across parameter draws as the attribution score for $\mathbf{x}_i^{(m)}$. We use the resulting scalar as the attribution score between the two sample-modality pairs.

LESS~\cite{xia2024less} estimates influence using optimizer-aware gradient similarity. After a short LoRA warm-up, it records model checkpoints and Adam optimizer states. For the masked candidate and target inputs, LESS computes an Adam-aware update feature for the candidate and a loss gradient for the target, projects both into a shared low-dimensional space, and accumulates their learning-rate-weighted cosine similarity across checkpoints. We use the resulting value as the attribution score between the two sample-modality pairs.

\subsection{Cross-sample Modality Interaction}

\begin{figure}[t]
    \centering
    \includegraphics[width=1.00\linewidth]{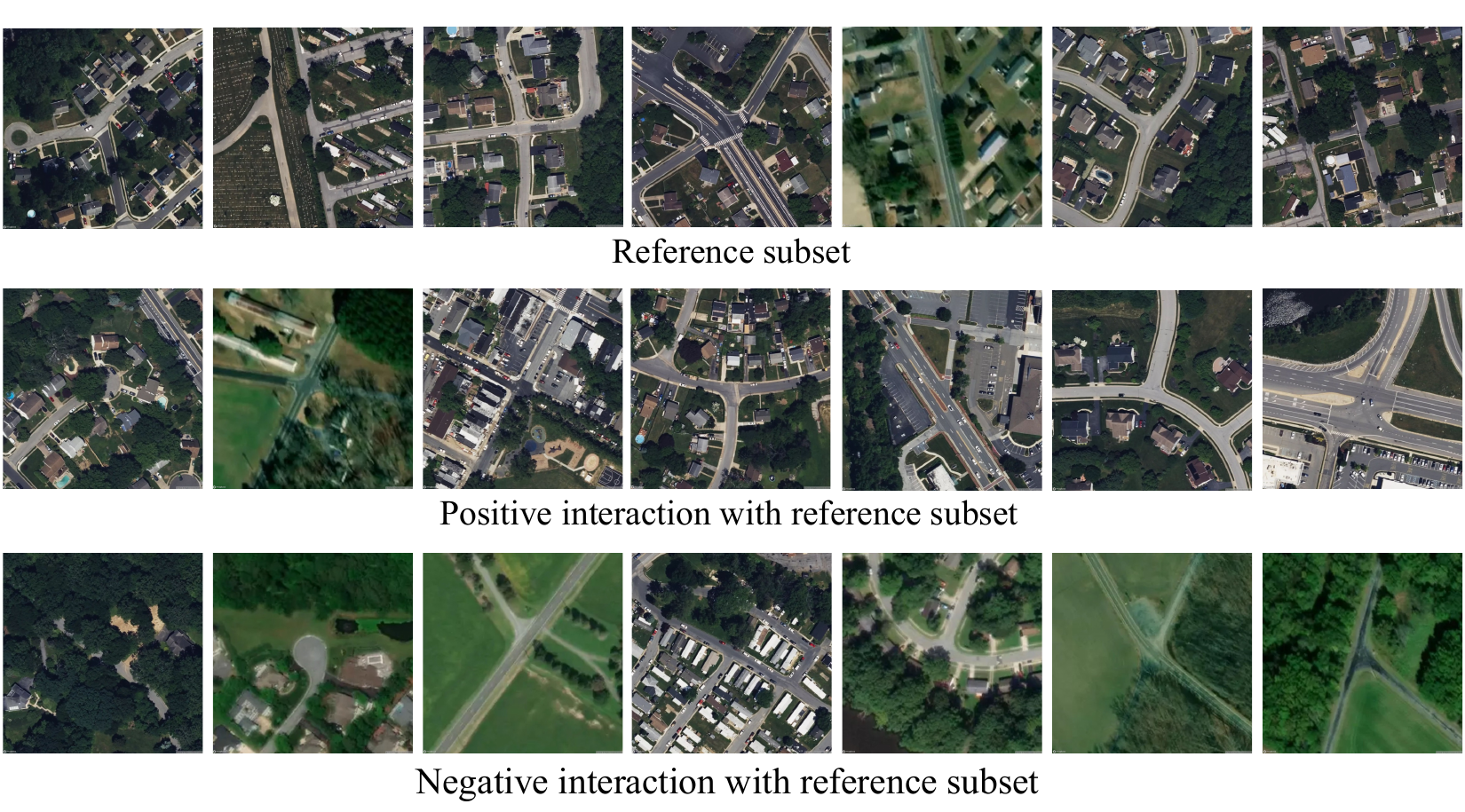}
    \caption{Examples of cross-sample modality interactions on MMTraCE. The top row shows the reference subset, while the middle and bottom rows show satellite images with positive and negative interactions with the reference subset, respectively. Positive interactions generally share similar road structures, and negative interactions exhibit substantially different spatial patterns.}
    \label{fig_interaction}
\end{figure}

We examine pairs of satellite images from MMTraCE to understand how visual information relates to interactions between samples.  As shown in Figure~\ref{fig_interaction}, positive interactions are observed when both images contain similar complex road structures, such as dense residential networks, curved or wide roads, and intersections. This shows that they reinforce each other. In contrast, negative interactions occur when the images differ substantially, such as when a dense urban scene is paired with a sparse, low-density, or rural scene. This demonstrates that the latter may weaken the evidence provided by the former. 

\subsection{First-order Approximation in Multimodal Models}\label{app_first_order}

To evaluate the accuracy of the first-order approximation, we measure the relative mean squared error (MSE) between the output of the real trained model and the approximated output using the validation set:
\begin{equation}
\operatorname{RMSE}(S) = \frac{ \sum_{j\in\mathcal{D}_{\mathrm{val}}} \left\| f_{W}(\mathbf{x}_j;S_{:,j}) - f_{W^\star}(\mathbf{x}_j;S_{:,j}) - \nabla f_{W^\star}(\mathbf{x}_j;S_{:,j})(W-W^\star) \right\|_2^2 }{ \sum_{j\in\mathcal{D}_{\mathrm{val}}} \left\| f_{W}(\mathbf{x}_j;S_{:,j}) \right\|_2^2 }.
\label{eq_multimodal_rrss}
\end{equation}

Table~\ref{tab_first_order_mmtrace} reports the approximation error and computational savings on \textsc{MMTraCE}, with computational cost measured in FLOPs. Across all six states, the approximation error remains lower than $1.2\%$. The linearized objective provides a speedup up to $18.9\times$ over exact retraining. These results show that the first-order approximation closely estimates the loss at a substantially lower computational cost.

\begin{table}[h]
\centering
\caption{First-order approximation error and speedup over exact retraining on MMTraCE. Errors are reported as the mean and standard deviation over three runs.}\label{tab_first_order_mmtrace}
\resizebox{0.8\columnwidth}{!}{
\begin{tabular}{lcccccc}
\toprule
 & DE & MA & MD & NV & MT & IA \\
\midrule
Relative MSE & $1.1_{\pm 0.2}\%$ & $1.0_{\pm 0.4}\%$ & $1.2_{\pm 0.1}\%$ & $0.8_{\pm 0.2}\%$ & $1.1_{\pm 0.1}\%$ & $1.1_{\pm0.5}\%$ \\
Speed up & $15.39\times$ & $4.07\times$ & $4.74\times$ & $5.62\times$ & $18.9\times$ & $14.7\times$ \\
\bottomrule
\end{tabular}}
\end{table}

The complexity of first-order approximation is linear in the size of the model, thus, we apply it to pretrained VLMs to show the scalability. We evaluate the accuracy of this estimation procedure on the multimodal ScienceQA benchmark~\cite{lu2022learn}, with model architectures ranging from 2B to 30B parameters. As shown in Table~\ref{tab_first_order_vlm}, the error is less than $1.1\%$ on all models.

\begin{table}[h]
\centering
\caption{First-order approximation error on ScienceQA across pretrained vision-language models ranging from 2B to 30B parameters. Results report the mean and standard deviation of the relative MSE over three runs.}\label{tab_first_order_vlm}
\resizebox{0.9\columnwidth}{!}{
\begin{tabular}{lcccccc}
\toprule
 & Qwen2-VL-2B & Qwen3-VL-8B & Qwen3-VL-30B-A3B  & Llama-3.2-11B-Vision \\
\midrule
Relative MSE & $3.6_{\pm 0.5}\times 10^{-3}$ & $2.9_{\pm 0.4}\times 10^{-3}$ & $1.1_{\pm 0.76}\times 10^{-2}$ & $1.1_{\pm 0.2}\times 10^{-2}$ \\
\bottomrule
\end{tabular}}
\end{table}

\subsection{Computational Costs}\label{app_flops_baseline}

By using the first-order approximation, \algo{} avoids retraining the multimodal model for each sampled state, allowing the cost of constructing state-loss pairs to scale linearly with the input size. We report the total computational cost, measured in FLOPs, on \textsc{MMTraCE} in Table~\ref{tab_flops_baseline}. Overall, \algo{} has a FLOP cost comparable to the baselines and is nearly identical to TracIn and LESS across all six tasks.

\begin{table}[h]
\centering
\caption{Computational cost of \algo{} and baseline attribution methods on \textsc{MMTraCE}, measured by the total number of FLOPs.}\label{tab_flops_baseline}
\resizebox{0.95\columnwidth}{!}{
\begin{tabular}{lcccccc}
\toprule
Method & DE & MA & MD & NV & MT & IA \\
\midrule
Influence functions & $6.09\times 10^{12}$ & $4.36\times 10^{14}$ & $1.49 \times 10^{14}$ & $5.86\times 10^{13}$ & $1.80\times 10^{13}$ & $7.75\times 10^{13}$ \\
DSIR & $4.28\times 10^{12}$ & $3.16\times 10^{14}$ & $9.23 \times 10^{13}$ & $3.81\times 10^{13}$ & $9.70\times 10^{12}$ & $4.85\times 10^{13}$ \\
TracIn & $5.26\times 10^{12}$ & $4.34\times 10^{14}$ & $1.47 \times 10^{14}$ & $5.65\times 10^{13}$ & $1.60\times 10^{13}$ & $7.54\times 10^{13}$ \\
LESS & $5.17\times 10^{12}$ & $4.34\times 10^{14}$ & $1.47 \times 10^{14}$ & $5.65\times 10^{13}$ & $1.60\times 10^{13}$ & $7.54\times 10^{13}$ \\
BIF & $5.73\times 10^{12}$ & $4.89\times 10^{14}$ & $1.64\times 10^{14}$ & $6.38\times 10^{13}$ & $1.68\times 10^{13}$ & $8.52\times 10^{13}$ \\\midrule
\algo{} & $5.15\times 10^{12}$ & $4.34\times 10^{14}$ & $1.47\times 10^{14}$ & $5.67\times 10^{13}$ & $1.59\times 10^{13}$ & $7.55\times 10^{13}$ \\
\bottomrule
\end{tabular}}
\end{table}

\subsection{Omitted Ablation Analysis}\label{app_kernel_choice}

\textbf{Choice of the RBF Kernel.}
We use the RBF kernel because it provides expressivity for any attribution function theoretically. \citet{zhang2026efficient} show that the Gaussian RBF kernel is strictly positive definite on any finite set of distinct vectors in $\mathbb{R}^d$. This implies that its Gram matrix is of full rank. Therefore, for any function $F$, there exists a kernel mapping that matches $F$ on the input vectors. Classical results \cite{poggio1990networks} guarantee that on this discrete domain, the RBF kernel can approximate any attribution function without loss of expressivity \cite{zhang2026efficient}.

Moreover, the RBF kernel encodes an inductive bias that matches the geometry of the subset space. On $\{0,1\}^d$, two subsets that differ on only a few coordinates are close in Hamming distance, and intuitively their performances tend to be similar. The RBF kernel directly reflects this structure. By contrast, polynomial kernels do not encode distance-based locality; two subsets with small Hamming distance are not necessarily mapped to nearby points in the feature space.

While we utilize the squared Euclidean distance function within the RBF kernel, the distance measure can be set to other types of measures depending on the application scenario. We conducted an ablation analysis comparing the RBF kernel to using an $\ell_2$-Laplacian kernel ($\exp(-\gamma \|S-S’\|_2)$), and another quadratic kernel. The results are reported in Table~\ref{tab_kernel_ablation}, which show that the RBF kernel gives a $31.8\%$ average advantage over the other two kernels in LDS scores on the MMTraCE benchmark.

\begin{table}[h]
\centering
\caption{Ablation study of nonlinear kernel choices on MMTraCE. Each entry reports the mean and standard deviation of the LDS score over three runs.}\label{tab_kernel_ablation}
\resizebox{0.83\columnwidth}{!}{
\begin{tabular}{lcccccc}
\toprule
Kernel & DE & MA & MD & NV & MT & IA \\
\midrule
$\ell_2$-Laplacian & $0.58_{\pm 0.03}$ & $0.55_{\pm 0.02}$ & $0.50_{\pm 0.02}$ & $0.48_{\pm 0.01}$ & $0.29_{\pm 0.03}$ & $0.65_{\pm 0.04}$ \\
Quadratic & $0.50_{\pm 0.02}$ & $0.47_{\pm 0.03}$ & $0.52_{\pm 0.01}$ & $0.40_{\pm 0.04}$ & $0.33_{\pm 0.01}$ & $0.56_{\pm 0.02}$ \\
RBF & $0.64_{\pm 0.03}$ & $0.62_{\pm 0.05}$ & $0.59_{\pm 0.10}$ & $0.58_{\pm 0.02}$ & $0.57_{\pm 0.06}$ & $0.70_{\pm 0.03}$ \\
\bottomrule
\end{tabular}}
\end{table}

\begin{figure}[t]
    \centering
    \begin{subfigure}{0.47\linewidth}
        \centering
        \includegraphics[width=0.6\linewidth]{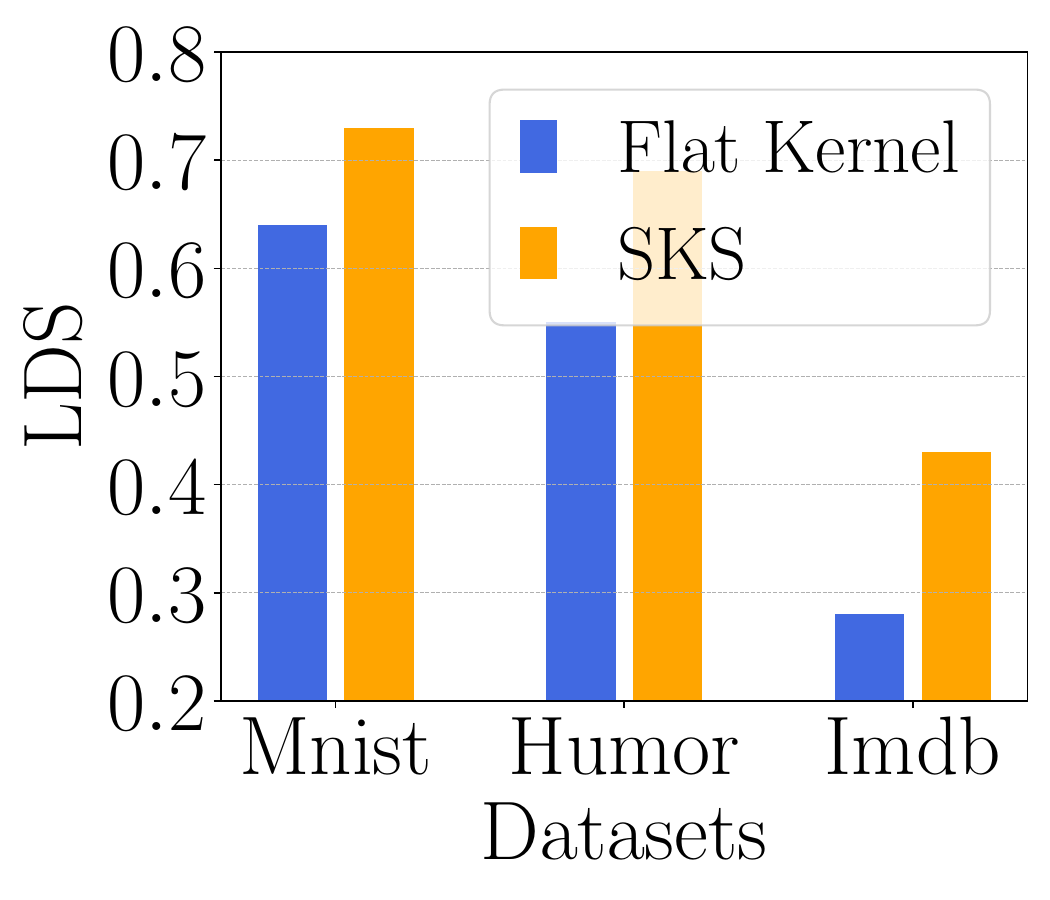}
        \caption{Compositional kernel}
        \label{fig_surrogate_omitted}
    \end{subfigure}\hfill
    \begin{subfigure}{0.47\linewidth}
        \centering
        \includegraphics[width=0.6\linewidth]{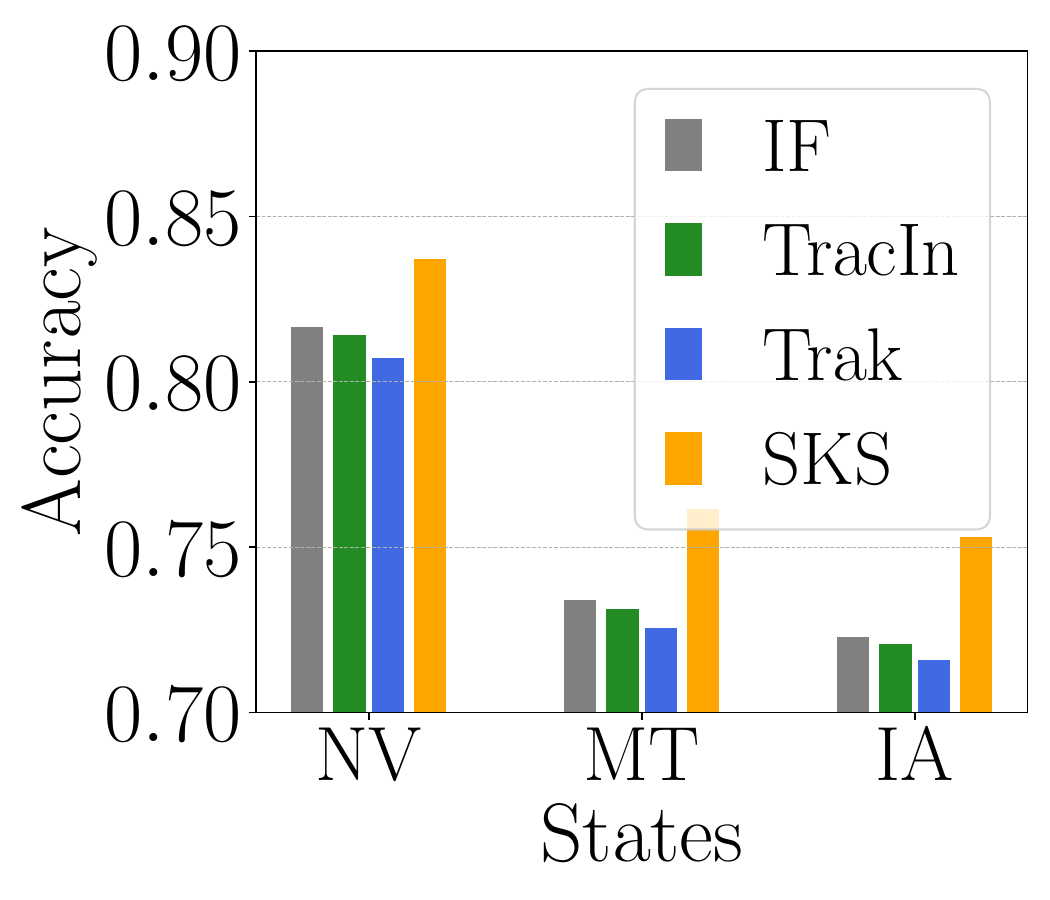}
        \caption{Data modality}
        \label{fig_modality_omitted}
    \end{subfigure}
    \caption{Ablation analysis on MultiBench and \textsc{MMTraCE}. Figure~\ref{fig_surrogate_omitted} compares the LDS of \algo{} with that of the flat surrogate kernel on AV-MNIST, Humor, and MM-IMDb. Figure~\ref{fig_modality_omitted} studies tabular-data selection on Nevada, Montana, and Iowa. We retain all visual data, select the top $20\%$ of the tabular data using each method, and report the resulting test accuracy.
    }
\end{figure}

\paragraph{Comparing the compositional kernel to a flat kernel design.} For the kernel-structure ablation analysis, the flat kernel vectorizes each sample-modality selection matrix $S$ into a single binary vector and applies one RBF kernel to the entire vector. It treats all sample-modality decisions as flat coordinates and does not distinguish sample-level similarity from modality-level similarity. We fit the same kernel ridge regression surrogate using each kernel and evaluate its LDS. Figure~\ref{fig_surrogate_omitted} reports the comparison on AV-MNIST, Humor, and MM-IMDb.

Next, we change which modality is selected under the fixed data budget. In the ablation, we retain all visual data and select the top $20\%$ of the tabular data according to each attribution method. We train the multimodal model on the resulting data subset and report its test accuracy. Figure~\ref{fig_modality_omitted} presents the remaining results on Nevada, Montana, and Iowa.

\subsection{Discussions}
\label{sec_discussions}
Our current empirical evaluation focuses on standard multimodal fusion models. This establishes a baseline for the efficacy of SKS but bounds our current scope. Investigating the scalability and behavior of SKS on large-scale Vision-Language Models (VLMs) is a promising direction for future research.